\pdfoutput=1

\documentclass[11pt]{article}

\usepackage[]{emnlp2021}

\usepackage[a-1b]{pdfx}
\usepackage{times}
\usepackage{amsmath}
\usepackage{amssymb}
\usepackage{graphicx}
\usepackage{caption}
\usepackage{multirow}
\usepackage{enumitem}
\usepackage{subcaption}
\usepackage{soul}
\usepackage{xcolor}
\usepackage{float}
\usepackage{booktabs}
\usepackage{natbib}
\usepackage{graphicx}
\usepackage{listings}
\usepackage{latexsym}
\usepackage{xcolor}
\usepackage[textsize=scriptsize]{todonotes}
\usepackage{algorithmicx}
\usepackage{hyperref}
\usepackage[ruled,vlined]{algorithm2e}
\usepackage{xr}
\usepackage[textsize=scriptsize]{todonotes}

\DeclareMathOperator*{\argmax}{arg\,max}

\definecolor{sig}{RGB}{194, 255, 200}

\usepackage[T1]{fontenc}

\usepackage[utf8]{inputenc}

\usepackage{microtype}

\usepackage[skins,breakable]{tcolorbox}
\definecolor{light-purple}{RGB}{151,156,171}
\definecolor{blue-color}{RGB}{40,166,189}
\definecolor{pink-color}{RGB}{237,46,104} 
\definecolor{dark-grey-color}{RGB}{79,91,102}

\newcommand{\promptsubsection}[1]{
\setlength{\parskip}{6pt} \noindent\textbf{{#1}:}
}

\newtcolorbox[list inside=prompt,auto counter,number within=section]{prompt}[1][]{
    colbacktitle=black!80,
    colframe=black!80,
    coltitle=white,
    fontupper=\footnotesize,
    boxsep=5pt,
    left=0pt,
    right=0pt,
    top=0pt,
    bottom=0pt,
    boxrule=1pt,
    enhanced, 
    breakable,
    skin first=enhanced,
    skin middle=enhanced,
    skin last=enhanced,
    #1,
}

\title{Contrastive Learning to Improve Retrieval for Real-world Fact Checking}

\author{Aniruddh Sriram\ \ \ \ \ \ \ \ \ \ \ Fangyuan Xu\ \ \ \ \ \ \ \ \ \ \ Eunsol Choi \ \ \ \ \ \ \ \ \ \ \ Greg Durrett \\
  Department of Computer Science \\
  The University of Texas at Austin \\
  \texttt{aniruddh.sriram@utexas.edu}}

\begin{document}
\maketitle
\begin{abstract}
Recent work on fact-checking addresses a realistic setting where models incorporate evidence retrieved from the web to decide the veracity of claims. A bottleneck in this pipeline is in retrieving relevant evidence: traditional methods may surface documents directly related to a claim, but fact-checking complex claims requires more inferences. For instance, a document about how a vaccine was developed is relevant to addressing claims about what it might contain, even if it does not address them directly. We present Contrastive Fact-Checking Reranker (CFR), an improved retriever for this setting. By leveraging the AVeriTeC dataset, which annotates subquestions for claims with human written answers from evidence documents, we fine-tune Contriever with a contrastive objective based on multiple training signals, including distillation from GPT-4, evaluating subquestion answers, and gold labels in the dataset. We evaluate our model on both retrieval and end-to-end veracity judgments about claims. On the AVeriTeC dataset, we find a 6\% improvement in veracity classification accuracy. We also show our gains can be transferred to FEVER, ClaimDecomp, HotpotQA, and a synthetic dataset requiring retrievers to make inferences.
\end{abstract}

\begin{figure}[t!]
\centering
\includegraphics[width=0.5\textwidth,trim=0mm 15mm 10mm 30mm]{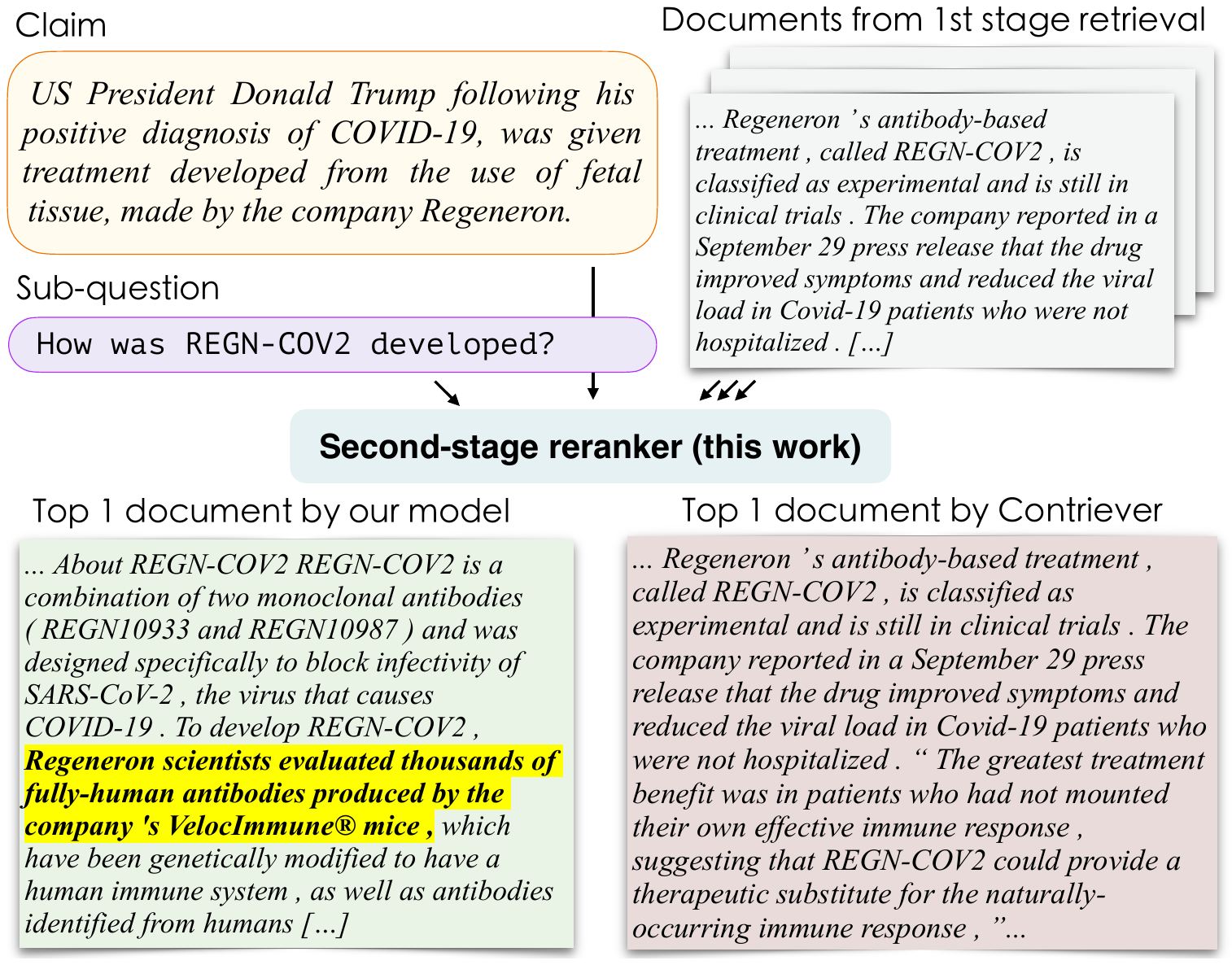}
\caption{Top-1 retrieved document from base Contriever (red) and CFR (green). Our model is able to choose a better document despite both paragraphs being topical.
Our model recognizes the question is asking about the chemical composition of REGN-COV2, while the unfinetuned model selects a relevant document that does not address ``fetal tissue'' or help with a final veracity judgment.
}
    \label{fig:motivating}
\end{figure}

\section{Introduction}

Retrieval-augmented generation (RAG) systems are now widely used across NLP applications including question answering \cite{realm, rag, karpukhin-etal-2020-dense} and text generation \cite{komeili-etal-2022-internet,Gao2023RetrievalAugmentedGF}, but one particular application of interest is fact-checking. While older fact-checking systems would often not consider evidence at all \cite{alhindi-etal-2018-evidence} or consider oracle evidence \cite{atanasova-etal-2020-generating-fact}, the real fact-checking task involves finding evidence to support or refute complex claims in the wild \cite{chen-etal-2022-generating, schlichtkrull2023averitec,chen2023complex}. As with many other RAG settings, retrieval is a bottleneck \cite{singh2022case}: it is impossible to provide the right judgment without retrieving the right evidence.

In this work, we investigate how to build an effective retriever for fact-checking. Figure~\ref{fig:motivating} shows an example of why this is particularly challenging: unlike a factoid question with a definite answer spelled out in text, documents retrieved for fact-checking may only obliquely address a  claim, or may present information in a different context (e.g., statistics that apply to a different country than the one where the claim was made). The unstructured nature of documents in the wild combined with claims that are only subtly true or false make retrieval a very difficult task.


We focus on two-step retrieval pipeline used in past work \cite{Lazaridou2022InternetaugmentedLM, chen2023complex}. These use a first-stage web search (i.e., using Google or Bing) to build a set of approximately relevant documents, followed by a second-stage fine-grained ranking to obtain a smaller set of documents to pass to a reader LM \cite{chen2023complex}, which produces the final veracity judgment. This second stage shows consistent recall failures despite high-quality documents being present in the first stage, mainly due to the nuanced complexities with claims and subquestions in fact-checking.

Our approach, Contrastive Fact-Checking Reranker (CFR), leverages contrastive learning to fine-tune a dense retriever to prefer more relevant documents when there is a lack of information or ambiguity in the claim. To train our model, we experiment with two main supervision signals: distilling knowledge from GPT-4 and measuring answer equivalence with the gold answer using Learned Equivalence Metric for Reading Comprehension (LERC) \cite{lerc}. We generate training datasets of positive and negative evidence pairs based on these signals and fine-tune Contriever \cite{izacard2022unsupervised}. 

Our evaluation shows that a combination of these supervision signals provides the best training data for the retriever, even better than fine-tuning on human annotated gold documents, as shown by gains in downstream performance across multiple datasets. 
Specifically, we see a 6\% improvement in veracity classification accuracy and a 9\% increase in the proportion of relevant top documents on AVeriTeC.

Our contributions are: (1) exploring new methods of supervision signals for contrastively training dense retrievers; (2) producing a strong dense retriever (CFR) which works well on AVeriTeC and a broader set of retrieval tasks regarding fact-checking complex claims. 




\section{Background and Related Work}

\subsection{Retrieval Augmented Generation Systems}

Retrieval-augmented generation (RAG) relies on two key modules: a retriever and a reader/generation model. For many RAG systems, noisy retrieval hurts downstream performance by providing irrelevant or misleading documents \cite{yoran2024making}. \citet{Sauchuk2022OnTR} found that adding distractors can cause a 27\% drop on veracity classification accuracy on FEVER. Therefore, it's important for retrievers to find relevant documents and simultaneously avoid damaging ones. \citet{Shi2023REPLUGRB} attempts to solve this problem by finetuning the retrieval component while fixing the reader LM, similar to our work. Other approaches like \citet{Ke2024BridgingTP} create a more complex system with a ``bridging'' model between the retriever and reader. Nevertheless, noisy retrieval remains a failure point in RAG systems \cite{Barnett2024SevenFP}, and tangible downstream gains can be realized by further finetuning.

\subsection{Limitations of Existing Retrieval Systems}

For NLP tasks like question answering, sparse retrieval techniques like BM25 have been supplanted by dense retrievers like DPR \cite{karpukhin-etal-2020-dense} and Contriever \cite{izacard2022unsupervised}. These dual encoder approaches support efficient retrieval, and contrastive training is an effective way to learn embeddings for QA tasks. More recently, research has explored distilling knowledge from reader models to create smarter retrievers \cite{izacard2022distilling}. We draw from this work to build a retrieval system with better reasoning capabilities than baseline dense retrievers, which are usually pretrained on simpler (query, document) pairs (i.e. the MSMARCO dataset). 
These retrieval systems have proven effective for fact-checking settings such as FEVER \cite{thorne-etal-2018-fever} 
and MultiFC \cite{augenstein-etal-2019-multifc}. 
However, the claims are largely short and factoid, and most of them contain no more than two entities. The realistic setting is embodied by approaches like QABriefs \cite{fan-etal-2020-generating}, ClaimDecomp \cite{chen-etal-2022-generating,chen2023complex}, and AVeriTeC \cite{schlichtkrull2023averitec}, 
which are ultimately different from what dense retrievers were developed and optimized for. 

\subsection{Motivating Example: AVeriTeC} 
Figure~\ref{fig:motivating} shows an example of fact-checking in the AVeriTeC dataset: \emph{``how was REGN-COV2 developed?''}. This example differs in key ways from frequently-studied question answering settings such as such as Natural Questions \cite{natquestions}.
First, it supports several different short answers but very likely has a best answer in the context of the claim: did the development involve human fetal tissue? In this case, the bolded paragraph indicates no: it used mice.
The answer to this question should address the claim and provide background information: there is both a ``short answer'' as well as a ``long answer'' \cite{natquestions, gao-etal-2023-enabling}. 



\paragraph{Retrieval signals in fact-checking} Contrastive methods like Contriever require examples marked as positive or negative for use in the contrastive objective.
In settings like NQ, retrieval systems rely on evaluating whether a retrieved passage contains the answer by simple string matching or ROUGE overlap, which identifies ``positives'' for retrieval. 
However, in Section \ref{results} we show it is not straighforward to apply this approach in fact-checking; i.e., we cannot simply say a passage is positive if it contains the ground truth answer.


Simultaneously, we must be cautious of assuming a low overlap with the answer indicates a ``negative'' document for retrieval. This is because multiple plausible answers can exist due to the open-ended nature of subquestions in AVeriTeC. Furthermore, using documents from the wild exacerbates this issue by introducing documents that might not directly support the gold answer but still contain valuable information about the claim. In Section \ref{Methodology}, we outline some ways in which we tackle this problem to curate better finetuning data.


\paragraph{Context in retrieval} 

Traditionally, retrievers are given standalone questions as queries. This is characteristic of datasets like NQ, where questions often contain one clear answer (e.g. ``\emph{Where is the bowling ball hall of fame located?}''). However, in fact-checking, the complexity of claims gives rise to subquestions that are not standalone or simple. Even if the questions themselves seem short (i.e., ``\emph{How was REGN-COV2 developed?}''), they must be interpreted in-context with the claim (i.e., ``\emph{Does REGN-COV2 contain fetal tissue?}''). Ideally, decomposing claims into a set of perfect standalone subquestions would reduce the load on the retriever. However, this itself is a hard and separate task. In this work, we attempt to build a retrieval system that can handle nuanced queries by considering each subquestion in the context of the overall claim.


\section{Methodology \label{Methodology}}

We consider a setting following work in AVeriTeC and ClaimDecomp \cite{chen-etal-2022-generating}. We assume we are given a collection of \textbf{claims} $(c_1,\ldots,c_N)$. For claim $c_i$, we define $q_{ij}$ as the $j$th \textbf{subquestion} for the $i$th claim in the dataset and $a_{ij}^g$ define its \textbf{answer}. We also assume access to a document set $D(c_i, q_{ij})$ for each subquestion, created by querying Bing with $c_i$ appended to $q_{ij}$ and scrape the top-k articles to form a document corpus.
Each \textbf{document} $d$ is a 200 token span gathered from the scraped articles. The title of the document is prepended to the start of each document. The dataset also comes with a \textbf{gold article} which contains the gold answer. Like the Bing-retrieved documents, it is chunked into 200 token span documents $\{d^g\}$ and added to $D(c_i, q_{ij})$. We refer to documents belonging to these articles as \emph{gold}. 

Given a query $y = [c_i; q_{ij}]$ and a document $d_i \in D$, we want to generate embeddings in $\mathbb{R}^e$ using an encoder network (e.g. Contriever). Let $h_y, h_{d_i}$ denote the representations of $y$ and $d_i$. Then we define our scoring function $f: \mathbb{R}^e \times \mathbb{R}^e \rightarrow \mathbb{R}$ such that $f(h_y, h_{d_i}) > f(h_y, h_{d_j})$ if document $d_i$ contains more information helpful to answering the query than document $d_j$.
Let $r(y) = \argmax_{d \in D}{f(h_y, h_d)}$ which is a function that chooses the highest ranked document in our document set $D$. 
The goal is to optimize our encoder via $f$ to rank  documents for answering questions in-context with the claim above topically relevant documents that do not ultimately contain information for an answer. We choose to optimize this for downstream veracity classification accuracy. We also track more upstream metrics such as using a relevance score for the top document or measuring how close its extracted answer matches the gold answer.

\begin{figure*}[t!]
  \centering
  \includegraphics[scale=0.65,trim=0mm 40mm 10mm 30mm]{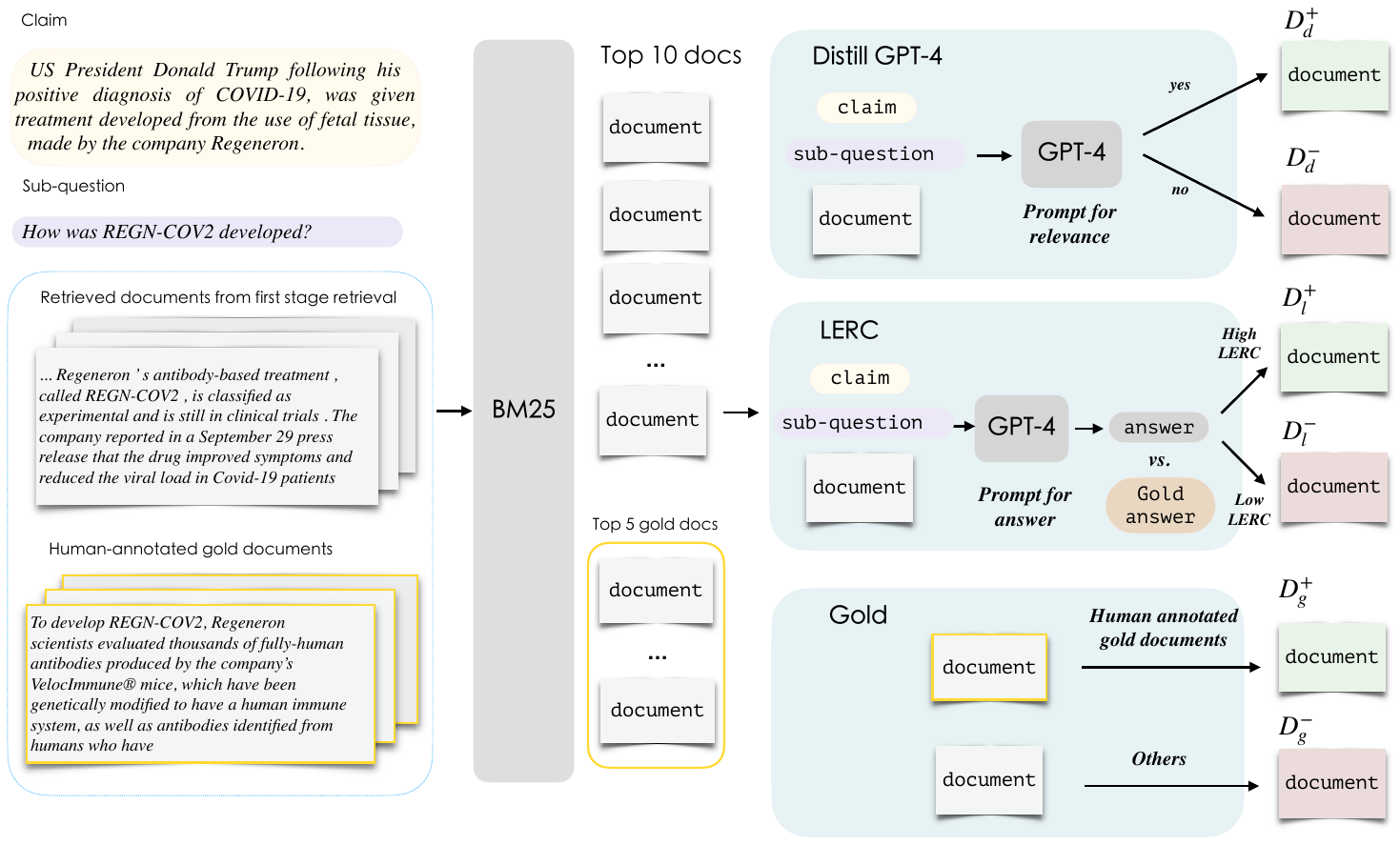}
  \caption{Overview of generating positive and negative examples for finetuning the retriever. We first select documents with high BM25 score with the (query, subquestion) from both the web documents and gold articles. We then experiment with different methods (described in Section \ref{subsec:training_data_generation}) to derive positive and hard negative examples.}
  \label{fig:training}
\end{figure*}

\subsection{Components}

\paragraph{Dense retriever $r$} We use Contriever as the base for our second stage dense retriever. Contriever uses the BERT base uncased architecture \cite{devlin-etal-2019-bert}. To fine-tune it with contrastive learning, we require document sets $T(c_i, q_{ij}, D) = \{D^+, D^{-}\}$ of positive and negative documents; during optimization, the positive documents will be embedded closer to the query vector than negative documents. Contrastive training relies critically on having hard negatives to serve as ``distractors'' \cite{robinson2021contrastive}. These might be documents ranked high by baseline retrievers or having high token overlap with the query. 
Figure~\ref{fig:training} shows our pipeline for constructing these document sets, which we expand on in the following sections.


We define $S_\text{BM25}(c_i, q_{ij}) = \{d_1, d_2, \dots, d_{k}\}$ as the top $k$ documents surfaced by BM25 given $[c_i; q_{ij}]$ as the query. We also define $G_\text{BM25}(c_i, q_{ij}) = \{d^g_1, d^g_2, \dots, d^g_l\}$ as the top $l$ gold annotated documents.
In our models, we set $k = 10$ and $l = 5$.


\paragraph{Reader model} We use GPT-4 as the reader model. The answers are derived by prompting GPT-4 with the claim $c_i$, question $q_{ij}$, and a document $d_{ij}$ from the corpus (see Appendix \ref{gptansprompt}).  For a given $(c_i, q_{ij})$ pair, we refer to $a_{ij}$ as the candidate answer derived from the evidence document $d_{ij}$. During inference time, $d_{ij}$ is the top-1 document from our retrieval system.

\subsection{Learning}

We train $r$ on these $(c_i, q_{ij}) \times T$ pairs to produce a finetuned retriever $r^*$. Specifically, given a query $y = [c_i; q_{ij}]$ and positive document $d^+ \in D^+$,

{
\scriptsize
\begin{align*}
    L(y, d^+) &= \frac{\exp\left(\frac{1}{\tau} f(h_y, h_{d^+}) \right)}{\exp\left(\frac{1}{\tau} f(h_y, h_{d^+} \right) + \sum_{d^- \in D^-} \exp\left(\frac{1}{\tau} f(h_y, h_{d^-}) \right)}
\end{align*}
}

\noindent
where $\tau$ is a temperature parameter. In our setting, we define $f$ as cosine similarity $\frac{h_y^Th_d}{\|h_y\|\cdot \|h_d\|}$ between the embeddings. 
This encourages positive documents to have high similarity with the query while penalizing high scores for negative documents. 
Fine-tuning yields $r^*$ such that $r^*(y)$ contains a better answer to $q_{ij}$ in context with $c_i$ than $r(y)$. 


\paragraph{Implementation Details}
On average, each question $q_{ij}$ comes with about 500 documents to rank. Each document contains 200 token span, scraped from articles with a 100 token length stride. Details about training and model architecture can be found in Appendix \ref{section:appendix:implementation}.

\subsection{Generating Contrastive Training Data}\label{subsec:training_data_generation}

We generate $\{D^+, D^-\}$ in three main ways: the annotated AVeriTeC gold evidence, distilled relevance judgements from a GPT-4 reader module, and evaluating equivalence of the document-predicted answer with a gold answer. Figure~\ref{fig:training} shows the three approaches which we describe next.

\paragraph{AVeriTeC Gold Evidence}

The most straightforward approach to building positive examples is to use the human-annotated evidence paragraphs available in AVeriTeC. The gold articles (one per subquestion) were selected by human annotators in a two-stage annotation process, we refer the readers to their paper for details \cite{schlichtkrull2023averitec}. The annotators also provided answers for the subquestions, which consist of both extractive and abstractive answers. For each $q_{ij}$, this article is chunked into a set of documents $\{d_{ij}^g\}$ as described in Section \ref{Methodology}. Negative examples are all $d \in S_\text{BM25}(c_i, q_{ij})$ such that $d$ is not from a gold-annotated document. We denote the fine-tuning data derived from this method as $\{ D_{g}^+, D_{g}^- \}$.

\paragraph{Distilling GPT-4} The AVeriTeC gold evidence may have recall errors: there may be relevant documents that are not marked by annotators. An alternative is to use GPT-4 as a labeler, effectively distilling its knowledge (Figure~\ref{fig:training}, top right). In this setting, we take $S_\text{BM25}(c_i, q_{ij})$ and zero-shot prompt GPT-4 about whether each document is relevant to answering the subquestion or not. Note we do not provide the gold answer in the prompt, as we are simply interested in collecting documents with relevant information regardless of how well the underlying answer matches $a_{ij}^g$. Documents marked as relevant are added to $D^+$, and the rest are added to $D^-$. The exact prompt can be found in Appendix \ref{relevanceprompt}. We define this set as $\{ D_{d}^+, D_{d}^- \}$. 

\paragraph{Distilling GPT-4 (with gold)} In this setting, we inject the top-$l$ AVeriTeC gold documents $G_\text{BM25}(c_i, q_{ij})$ into the finetuning set. Like before, we zero-shot prompt GPT-4 about whether each document is relevant to answering the subquestion, but include $G_\text{BM25}(c_i, q_{ij})$ in addition to $S_\text{BM25}(c_i, q_{ij})$. We refer to $\{ D_{dg}^+, D_{dg}^- \}$ as the finetuning data from this method.

\paragraph{LERC-based signal \label{lercmethod}}
An additional approach to construct our pairs is to use the gold-annotated answers $a_{ij}^g$ (Figure~\ref{fig:training}, middle right). Ideally, a document we retrieve should help us discover these answers; however, because the subquestions are not factoid questions, it is not easy to assess whether a retrieved document contains the answer.

To do this, we filter the top documents using LERC (Learned Evaluation Metric for Reading Comprehension)~\cite{lerc}, a metric for scoring answer equivalence. More formally, we take $S_\text{BM25}(c_i, q_{ij})$ with $G_\text{BM25}(c_i, q_{ij})$ to make a set of 15 documents. We then prompt GPT-4 to use each of the 15 evidence documents to produce an answer $a_{ij}$ for each document. We found that for complex long answers, using ROUGE overlap as an answer equivalence metric works poorly (Appendix \ref{section:appendix:rougeoverlap}). On AVeriTeC, we also tried using ROUGE-F1 score instead of LERC (see Table \ref{tab:main_results}) to see how this reflects in all our end-to-end evaluation metrics. To accommodate this, we introduce an ``answer shortening'' function $s$ which attempts to pull out the main point of the answer. We use LERC to compare $s(a_{ij})$ and $s(a_{ij}^g)$, our shortened candidate and gold answer respectively. By identifying documents which give rise to answers with high LERC scores, we encourage our retriever to seek documents which address the question in the query. Documents with poor LERC scores ($< 0.3$) become negative contexts, and documents with high LERC ($>0.7$) scores are positive contexts. We also evaluate how well human annotators agree with granular LERC scores and find an average Kendall's $\tau$ score of 0.53 (Appendix \ref{section:appendix:lerc}). We denote $\{ D_{l}^+, D_{l}^- \}$ as finetuning data derived from this method.

\paragraph{LERC-based quality check} 
We evaluated $\{ D_{l}^+, D_{l}^- \}$ and found that many negative documents were actually relevant to the claim/question. More details on this experiment can be found in Appendix \ref{section:appendix:lercqualitycheck}. To reduce the false negative rate, we mix in relevant documents with the positive set from \emph{distill} to create $\{D_{dg}^+ \cup D_l^+, D_{dg}^-\}$. We refer to this as the \emph{distill (gold) + LERC} setting. This is the final experimental setting we use for our \textbf{Contrastive Fact-Checking Reranker (CFR)} model.

\begin{table}[t!]
\small
  \centering
  
  \begin{tabular}{@{}cccccc@{}}
    \toprule
    Train Set  & \# subq & $|D^+|$ & $|D^-|$ &  $D^+$ & $D^-$\\
    \midrule
    distill  & 1228 & 4.8 & 8.4 & $D_d^+$ & $D_d^-$ \\
    LERC  & 692 & 1 & 4.2 & $D_l^+$ & $D_l^-$\\
    gold & 1229 & 1 & 9.1 & $D_g^+$ & $D_g^-$\\
    distill (gold) & 1229 & 5.2 & 8.4 & $D_{dg}^+$ & $D_{dg}^-$\\
    distill (gold)\\ + LERC & 1229 & 5.6 & 8.4 & $D_{dg}^+ \cup D_l^+$ & $D_{dg}^-$ \\
    \bottomrule
  \end{tabular}
  \caption{Dataset statistics for different finetuning sets from AVeriTeC. $|D^+|$ and $|D^-|$ represent the average number of positive and negative contexts per $(c_i, q_{ij})$  pair. Differences in number of subquestions come from filtering out examples for which $|D^+| = 0$ or $|D^-| = 0$.
  }
\end{table}

\section{Experimental Setup}\label{sec:exp}


We evaluate Contriever fine-tuned on the supervision signals outlined in Section \ref{Methodology}. The datasets selected for evaluation, namely AVeriTeC \cite{schlichtkrull2023averitec}, ClaimDecomp \cite{chen-etal-2022-generating}, FEVER \cite{thorne-etal-2018-fever}, and HotpotQA \cite{yang-etal-2018-hotpotqa}, encompass a wide range of scenarios for document retrieval. 
For evaluation, a random subset of 200 answerable examples (subquestions contain an answer) were selected from each of these not overlapping with the training sets.



\subsection{Metrics}

We use metrics that evaluate both the retrieved documents and downstream products of these documents, such as the produced answer.
\begin{itemize}
\item \textbf{LERC} computes the average LERC score between the AVeriTeC (or ClaimDecomp) gold answer and the GPT-4 generated answer from the top retrieved document as the candidate.
    \item \textbf{Top doc relevance} is the proportion of examples for which the top-1 document is classified as relevant to answering the question by GPT-4, using the same prompt for which we derive the distillation signal.
\item \textbf{Gold@10} is the proportion of examples in which an AVeriTeC annotated gold document appeared in the top-10.

\item \textbf{Veracity} represents the veracity classification accuracy. For ClaimDecomp, we use the RoBERTa based veracity classifier trained on ClaimDecomp.\footnote{\url{https://github.com/jifan-chen/Fact-checking-via-Raw-Evidence}} For FEVER, we few-shot prompt GPT-4 for a veracity label; see Appendix \ref{feververacityprompt}. 
\end{itemize}



\subsection{Datasets}

\paragraph{AVeriTeC} consists of real claims ($c_i$) from the web annotated with subquestions ($q_{ij}$), gold answers ($a^g_{ij}$) to the subquestions, and the gold evidence document for the answer. We query Bing in FSR with the claim and subquestion $[c_i; q_{ij}]$ to generate $D$. The generated answers ($a_{ij}$) are verified against the gold answers using LERC.

\paragraph{ClaimDecomp} consists of complex political claims ($c_i$) with yes/no subquestion decompositions ($q_{ij}$) generated by trained annotators. We query Bing in FSR with the claim and subquestion $[c_i; q_{ij}]$ 
to generate $D$. The annotated subquestions tackle both explicit and implicit parts of the original claim. The implicit questions are much harder to answer without sufficient context, which makes this an interesting dataset for retrieval evaluation. The human labeled answers are yes/no, and we evaluate our generated answers ($a_{ij}$) against the gold answers using LERC. Because the questions themselves are yes/no in nature, this approach returns the same results as simple binary comparison.

\paragraph{FEVER} consists of claims $(c_i)$ manually verified against the introductory sections of Wikipedia pages and classified as \textsc{supported}, \textsc{refuted} or \textsc{notenoughinfo}. We treat the claim itself as the question ($c_i = q_{i}$) here. Unlike past work, we query Bing with the claim to generate $D$; as a result, our data condition is different than past work evaluating on FEVER. For FEVER, 
we don't generate answers or subquestions and simply verify the claim against the evidence document.

\paragraph{HotpotQA} is a question answering dataset featuring multi-hop questions, with strong supervision for supporting facts to enable more explainable question answering systems. The questions require finding and reasoning over multiple supporting documents to answer. There are no claims in this dataset, so we set $c_i = q_i$ and  retrieval is done with just the question. 

\subsection{Baselines}
We report performance of several widely-used retrievers as baselines: \textbf{BM25}, \textbf{Contriever} \cite{izacard2022unsupervised} and Contriever fine-tuned on the MS MARCO dataset \cite{Campos2016MSMA} (\textbf{Contriever MSM}). We also compare against an additional Contriever baseline. We use \textbf{ROUGE-F1} supervision similar to the LERC setup, except long answers were evaluated using ROUGE overlap scores. This tests whether our approaches outperform a simple method for answer matching.

\section{Results \label{results}}

    
\begin{table}[t]
\footnotesize
  \centering
  \renewcommand{\tabcolsep}{1.1mm}  
  \begin{tabular}{@{}cccccc@{}}
    \toprule
    Model & LERC & Top Doc Relv. & Gold@10 & Veracity \\
    \midrule
    BM25 & 0.45 & 0.47 & 0.42 & 0.48 \\
    Contriever & 0.48 & 0.54 & 0.50 & 0.54 \\
    Contriever MSM & 0.52 & 0.55 & 0.45 & \colorbox{sig}{0.59} \\
    ROUGE-F1* &0.52 & 0.53 & 0.50 & 0.55 \\
    \midrule
  gold & 0.50 & 0.51 & 0.56 & 0.53 \\
   distill & \colorbox{sig}{\textbf{0.54}} & \colorbox{sig}{\textbf{0.63}} & \colorbox{sig}{\textbf{0.60}} & 0.55 \\
   LERC & \colorbox{sig}{0.53} & 0.56 & 0.54 & \colorbox{sig}{\textbf{0.60}} \\
 distill (gold) & \colorbox{sig}{\textbf{0.54}} & \colorbox{sig}{0.61} & \colorbox{sig}{0.59} & \colorbox{sig}{0.58} \\
 CFR & \colorbox{sig}{0.53} & \colorbox{sig}{0.62} & \colorbox{sig}{0.59} & \colorbox{sig}{\textbf{0.60}} \\
    \bottomrule
  \end{tabular}
  \caption{In-domain experimental results on AVeriTeC test subset ($n=200$). Numbers marked with \colorbox{sig}{ } are statistically significant w.r.t.~baseline Contriever at $o = 0.05$ under 10,000 bootstrapped samples. CFR is what we call the model finetuned on \textit{distill (gold) + LERC}.}
  \label{tab:main_results}
\end{table}

\begin{table*}[t!]
\centering
\small
\begin{tabular}{@{}cccccccc@{}}
\toprule
    \multirow{2}{*}{Model} & \multicolumn{3}{c}{ClaimDecomp} & \multicolumn{2}{c}{FEVER} & \multicolumn{2}{c}{HotpotQA} \\
\cmidrule(lr){2-4} \cmidrule(lr){5-6} \cmidrule(lr){7-8}
& LERC & Top Doc Relv. & Veracity & Top Doc Relv. & Veracity & LERC & Top Doc Relv. \\
\midrule
BM25 & 0.54 & 0.30 & 0.30 & 0.43 & 0.55 & 0.28 & 0.21 \\
Contriever & 0.64 & 0.32 & 0.32 & 0.49 & 0.58 & 0.33 & 0.27 \\
Contriever MSM & 0.64 & 0.31 & 0.34 & 0.52 & 0.61 & 0.34 & 0.31 \\
\midrule
gold & 0.64 & 0.30 & 0.28 & 0.48 & 0.56 & 0.32 & 0.30 \\
distill & 0.64 & \colorbox{sig}{\textbf{0.39}} & 0.32 & \colorbox{sig}{\textbf{0.57}} & 0.61 & 0.34 & 0.26 \\
LERC & 0.65 & 0.31 & 0.31 & 0.55 & 0.61 & 0.34 & 0.30 \\
distill (gold) & \colorbox{sig}{\textbf{0.66}} & \colorbox{sig}{0.37} & 0.34 & 0.56 & 0.61 & 0.35 & \colorbox{sig}{\textbf{0.32}} \\
CFR & 0.65 & 0.32 & 0.34 & \colorbox{sig}{\textbf{0.57}} & \colorbox{sig}{\textbf{0.63}} & \textbf{0.36} & \colorbox{sig}{\textbf{0.32}} \\
\bottomrule
\end{tabular}
\caption{Out-of-domain experimental results on ClaimDecomp, FEVER, and HotpotQA test subset (n=200 for each dataset). Numbers marked with \colorbox{sig}{ } are statistically significant w.r.t. baseline Contriever at $p = 0.05$ under 10,000 bootstrapped samples from the respective test subset.}\label{tab:results_other_datasets}
\end{table*}

\subsection{AVeriTeC}

The results for AVeriTeC are shown in Table \ref{tab:main_results}. We find that \emph{distill} performs the best in most metrics but for veracity. The 6\% gain in top doc relevance reflect our retriever's ability to correctly identify more relevant documents in our evaluation set. 


As expected, we find that using ROUGE as a long answer overlap metric to generate $\{D^+, D^-\}$ works poorly as seen by the ROUGE-F1 baseline.

Comparing the average LERC score between baseline Contriever and Contriever finetuned on \textit{LERC}, we find a 5\% gain in the average LERC score on the evaluation set. This is also backed by a 6\% increase in downstream veracity classification performance, indicating our improved ability to answer questions transfers to actually fact-checking the claim. We also see that the models finetuned with LERC signals (\emph{LERC} and CFR) reflect the strongest improvements in veracity classification. CFR also excels in top doc relevance and other upstream metrics. This indicates evaluating answers derived from documents may help downstream performance on fact-checking more than other supervision signals.

\paragraph{Lexical overlap} We find that \emph{gold} supervision (using AVeriTeC annotated gold evidence) performs poorly across all metrics. 
We hypothesize two reasons for this: 1) the evidence lacks significant token overlap with the claim/subquestion and 2) gold annotation involves human reasoning and assumptions which are too complex for the unfinetuned retriever to model in its document embedding space. In fact, the average ROUGE-F1 score between $[c_i; q_{ij}]$ and highest overlapping gold document is only 0.11 compared to 0.25 for the top-ranked document from the wild (see Appendix \ref{section:appendix:tokenoverlap}). This discrepancy comes from examples where the annotated evidence document is based on a related entity not mentioned in the claim or question, which is very challenging to recover without additional context. In other cases, modeling the annotated gold evidence is challenging because it contains new information that is not known from the claim or subquestion alone. Therefore, supervising with only gold documents doesn't effectively help the retriever learn. 



\subsection{Out-of-domain results}
Results on out-of-domain datasets are in Table~\ref{tab:results_other_datasets}.

\paragraph{ClaimDecomp} We find that our gains translate to ClaimDecomp, with \textit{distill (gold)} demonstrating significant improvements in both LERC and top doc relevance. Examples in this dataset contains both explicit and implicit subquestions,
while AVeriTeC subquestions are mostly explicit. Since we use subquestions for retrieval, improvement in top doc relevance may reflect an ability to surface better documents for ambiguous implicit subquestions, which is something baseline retrievers struggle with. An example of this is seen in Appendix \ref{section:appendix:claimdecomp}, where our finetuned retriever model is able to accurately capture the focus on lack of funding presented in the question. Even though baseline Contriever selects a document detailing the Amtrak incident with high lexical overlap with the claim and query, the document itself is not useful for answering the question. Using CFR, we see a 2\% increase in downstream veracity classification performance.

\paragraph{FEVER} We also find that our system gives gains on FEVER compared to BM25, Contriever, and Contriever MSM. Our retriever selects relevant top documents more often and yields improved downstream veracity performance.

\paragraph{HotpotQA} For HotpotQA, we find that \emph{distill (gold) + LERC} performs the best across LERC and top doc relevance. We notice the strongest gains come from including LERC-based supervision, which indicates our retriever may learn to identify answer documents that contain little overlap with the claim. This is especially useful in multi-hop settings where the answer document cannot be found in one step from the query. 




\section{Retriever Reasoning Capabilities}

Our hypothesis about our contrastive training was that it would impart a greater ability for our retriever to ``reason'' about content rather than directly locating an answer. We conduct an additional study of whether our retriever can exhibit basic 1-hop reasoning capabilities via a synthetic data experiment. We construct positive and negative documents where the positive documents do not directly state the answer, similar to what we found in several AVeriTeC examples. 

\subsection{Synthetic Data Generation}

We build these examples by few-shot prompting GPT-4 with synthetic documents written by humans.  Our data generation approach takes as input a claim/question pair $(c_i, q_{ij})$ from AVeriTeC and produces a document set $\{d^+, d^-, d_1^-, d_2^-, d_3^-, d_4^-\}$. We generate data for  $(c_{i}, q_{ij})$ pairs from the validation set described in Section \ref{sec:exp}. The positive document $d^+$ is the only document that contains an answer to the question. Document $d^-$ is a ``hard negative'' document, which is a document that appears \textit{highly} relevant to the query $[c_i; q_{ij}]$ but does not contain an answer. The 4 other documents $d_1^-,\dots,d_4^-$ are additional negative documents built from alternate subquestions about the claim. 

The \textbf{positive document} is a paragraph that supports an answer to the question, but only indirectly. When prompting (Appendix \ref{syntheticprompt}), we require that a clear reasoning hop must be made to recover an answer from the positive document. Therefore, a retrieval system that simply looks for query-document token overlap may not be able to find such documents because the answer is usually not presented in terms of the question. 

The \textbf{hard negative document} is a paragraph that looks highly relevant to the claim/question, but doesn't actually support an answer. In the prompt, we specify that the document should appear relevant but not support an answer, and further enforce this with few-shot examples (see Appendix \ref{syntheticprompt}). In Appendix \ref{section:appendix:synthetichuman}, the hard negative document correctly discusses the federal judges Trump nominated. However, it does not contain any information about \textit{how many} judges he nominated, deeming it useless for answering the question about the claim.

The \textbf{remaining negative documents} are built by generating alternate subquestions similar to $q_{ij}$ but without overlapping answers. Then, we generate documents that contain answers to these distractor subquestions. An example can be found in Appendix \ref{section:appendix:syntheticgenerated} along with the prompt in Appendix \ref{syntheticprompt}.

\subsection{Results}
We evaluate our retrievers on their ability to score the positive document closer to the query than the negative distractor documents. We measure this via MRR of the positive document across ranking the six documents (positive, hard negative, and 4 alternate question negatives). The results are displayed in Table \ref{tab:synthetic_results}. We find a statistically significant gain in our finetuned model's ability to surface the positive document over other distractor documents. CFR achieves an MRR of 0.79 compared to baseline Contriever (0.68). This supports our hypothesis that finetuning on our supervision signals improves the ability of the retrieval model to find information only indirectly related to the claim.  

\begin{table}[t!]
\small
\footnotesize
  \centering
  \renewcommand{\tabcolsep}{1.1mm}  
  \begin{tabular}{@{}ccc@{}}
    \toprule
    {Model}  & MRR \\
    \midrule
    BM25 & 0.49\\
    Contriever & 0.68\\
    Contriever MSM & 0.75 \\
    \midrule
   gold & 0.72 \\
   distill & \colorbox{sig}{\textbf{0.80}} \\
   LERC & 0.72 \\
 distill (gold) & \colorbox{sig}{\textbf{0.80}} \\
 CFR & \colorbox{sig}{0.79} \\
    \bottomrule
  \end{tabular}
  \caption{Results for 200 examples of synthetically generated data. Numbers marked with \colorbox{sig}{ } are statistically significant w.r.t.~ baseline Contriever at $p = 0.10$ under 10,000 bootstrapped samples from the respective test set.}
  \label{tab:synthetic_results}
\end{table}

\section{Conclusion}
This work presents an improved retrieval system, CFR, for fact-checking complex claims. We present two supervision signals for finetuning retrievers under a contrastive objective, and their integration results in improved downstream veracity classification. Furthermore, CFR is able to improve retrieval in settings where inferences are required to identify the correct documents. The gains found in this paper encourage explorations into improving retrieval for fact-checking, as surfacing relevant information proved to be a hard task even for SOTA dense retrievers. 

\section*{Limitations}

There are a few limitations of our current approach. First, using LERC as an answer equivalence metric requires us to shorten both the gold and candidate answer. The answer compression step loses information that may play a role in verifying hard examples. Therefore, developing a good long answer equivalence metric can help build an even better retrieval system for fact-checking. Such equivalence metrics can also be useful for evaluation: the long-form explanation of why a claim is true or false may be more important than the veracity judgment itself, but this is difficult to assess in an automated way.

Second, this work focuses on the second-stage retrieval step. Building optimized queries for first stage retrieval may yield a better document corpus for second stage, especially for hard examples where little information has been published. However, indexing the necessary documents for the broad set of claims we use involves web-scale indexing, which is beyond the scope of this project.

Finally, this work considered English-language political claims. We note that claims in multimedia (e.g., in memes or videos), claims in other languages, and claims in specialized domains such as COVID-19 misinformation may present distinct challenges. However, we believe that our framework is flexible enough for future work to be able to build on it and train retrievers for these settings as well.

\section*{Ethical Considerations and Risks}

This paper presents a retrieval method that seeks to advance the state of the art in automated fact-checking. However, despite recent progress in this area and systems that combine retrieval systems like ours with LLMs \cite{schlichtkrull2023averitec,chen2023complex}, we stress that these system are not yet ready for deployment. We believe these systems have use to aid professional fact-checkers in their work, since enabling them to quickly find information can aid them to more rapidly check claims. However, these systems cannot produce reliable fact-checks without a human in the loop, as demonstrated by the veracity numbers in this work. Moreover, there is not necessarily a single objective truth about every claim, and a judgment may depend on the reliability of primary sources and other factors which are beyond the scope of this work. 

\section*{Acknowledgments}

This work was partially supported by Good Systems,\footnote{https://goodsystems.utexas.edu/} a UT Austin Grand Challenge to develop responsible AI technologies, NSF CAREER Award IIS-2145280, and the NSF AI Institute for Foundations of Machine Learning (IFML). We thank the UT Austin NLP community for revisions and feedback on earlier drafts of this paper.

\bibliography{anthology,custom}
\bibliographystyle{acl_natbib}

\newpage

\appendix


\section{Implementation Details}
\subsection{Computational Details}
\label{section:appendix:implementation}

The finetuned models were BERT base uncased (110M parameters). Hyperparameter optimization was done via grid search on the learning rate and batch size. For learning rate, we searched $\{1e-5, 2e-5, 4-e5\}$. For batch size, we searched $\{4, 8, 16, 32, 64\}$.
\begin{itemize}
    \item Infrastructure: 2 NVIDIA Quadro RTX 8000
    \item GPU Hours (training): approx. 3 hours
    \item GPU Hours (eval): approx. 1 hour
    \item Epochs: 12 
    \item Best Learning Rate: 2e-5
    \item Best Batch Size: 32
\end{itemize}

\subsection{Experimental Setup}
Besides chunking into 200 token spans, document text is not further preprocessed. During training, data was mapped into tuples of the form containing one positive and negative $(c_i, q_{ij}, d^+, d^-)$. That is, if a claim/question pair contains 2 positive and 3 negative paragraphs, it becomes $2\cdot3=6$ separate data points. These were then shuffled and batched to be fed to the retriever. In contrastive training we use in-batch negatives.

\subsection{Parameters for Packages}
\begin{itemize}
    \item Used rouge-score (v0.1.2) to compute ROUGE-F1 scores. Used \texttt{rougeL} (longest common subsequence) with stemming set to True.
    \item Used openai (v1.34.0) for GPT-4 chat completion. Set temperature setting to 0.2.
\end{itemize}

\subsection{Scientific Artifacts}
\begin{itemize}
    \item \textbf{AVeriTeC} \href{https://github.com/hotpotqa/hotpot/blob/master/LICENSE.txt}{[License]} Free to copy, redistribute, and build upon this material given citations and a link to the license. AVeriTeC contains English-language real-world claims mainly in politics gathered from 50 different fact-checking organizations.
    \item \textbf{FEVER} \href{https://fever.ai/download/fever/license.html}{[License]} Data annotations incorporate material from Wikipedia, which is licensed pursuant to the Wikipedia Copyright Policy
    \item \textbf{HotpotQA} \href{https://github.com/hotpotqa/hotpot/blob/master/LICENSE.txt}{[License]} Free to copy, redistribute, and build upon this material given citations and a link to the license
    \item \textbf{Contriever} \href{https://github.com/facebookresearch/contriever/blob/main/LICENSE}{[License]} Free to copy, redistribute, and build upon this material given citations and a link to the license
\end{itemize}

\section{ROUGE-based Methods}
\subsection{ROUGE-based Answer Matching}
\label{section:appendix:rougeoverlap}

ROUGE overlap between long answers works is a poor supervision signal because answer strings are typically quite complex. Table \ref{table:rouge} illustrates this: although both long answers are conveying the same fact that Nigeria experienced 29 years of military rule, extra details or differences in phrasing can lead to low ROUGE scores despite the answers being semantically equivalent. The opposite may also occur: long answers which contain high lexical overlap may be topically similar but completely different in their key points, creating a false positive example. We also investigated semantic similarity measures like BERT score to assess answer equivalence. Compared to short answer LERC, BERT score tended to work poorly for complex long answers as seen in AVeriTeC. By contrast, using a short answer extraction yields a perfect signal in this case.

\begin{table*}[h!]
\small
  \centering
  \begin{tabular}{p{0.2\textwidth}|p{0.3\textwidth}|p{0.3\textwidth}|p{0.1\textwidth}}
    \toprule
    & \textbf{Gold Answer} & \textbf{GPT-4 Answer} & \textbf{Score} \\
    \midrule
    Long Answer + ROUGE-F1 & Nigeria returned to democracy in 1999, after two long periods of military rule—1966–79 and 1983–98—during which the military wielded executive, legislative, and judicial power & Nigeria experienced military rule for a total of 29 years after independence: from 1966 to 1979 and from 1983 to 1998. & 0.22 \\
    \midrule
    Short Answer + LERC & 29 years & 29 years & 1 \\
    \bottomrule
  \end{tabular}
  \caption{Comparison of long answer ROUGE and short answer LERC. The two long answers are effectively conveying the same thing, but the ROUGE-F1 score is only 0.22. However, answer shortening + LERC yields a perfect equivalence score of 1.}
  \label{table:rouge}
\end{table*}

\subsection{ROUGE-based Token Overlap}
\label{section:appendix:tokenoverlap}

See Table~\ref{table:tokenoverlap}. The token overlap between the retriever query (claim+question) and the AVeriTeC annotated gold document is only 0.11, whereas with the top retrieved document it is 0.25. This means using tokens in the query to surface the gold document is not easy.

\begin{table}[h!]
  \centering
  \renewcommand{\tabcolsep}{1.1mm}  
  \begin{tabular}{lcc}
    \toprule
    & gold  & top\_doc \\
    \midrule
    ROUGE-F1 & 0.11 & 0.25 \\
    \bottomrule
  \end{tabular}
  \caption{Comparing token overlap across 200 examples between $[c_i; q_{ij}]$ and the best annotated gold document or the top-ranked document from the wild (retriever is baseline Contriever).}
  \label{table:tokenoverlap}
\end{table}

\section{LERC Experiments}

\subsection{LERC Quality Check}
\label{section:appendix:lercqualitycheck}
We evaluate the selection of $\{ D_{l}^+, D_{l}^- \}$ by manually annotating 10 examples. The task was to select the positive context document given a shuffled, unlabeled $\{ D_{l}^+, D_{l}^- \}$. We selected the positive document correctly in 60\% of examples. Note the positive document here is the one with the highest LERC score (i.e., contains an answer which most closely matches the gold answer). However, the two human annotators agreed on 90\% of examples. By investigating the failure cases, we found that LERC-based metrics are sensitive to selecting false negative documents, as human agreement indicated a negative document was more ``relevant'' to the claim/question than the labeled positive document 40\% of the time.
Oftentimes, the misclassified document contained a reasonable answer to the question but mismatched the gold answer (hence explaining the low LERC score). This revealed that while LERC can identify strong positive documents, it comes with the risk of including relevant documents as negative contexts. 

\subsection{LERC-Human Agreement}
\label{section:appendix:lerc}

In another preliminary study, we manually annotated 22 examples with a fine-grained score from 0-1 reflecting how closely we think the shortened candidate answer matches the shortened gold answer. Across three annotators, we found Kendall's tau agreement scores of 0.55, 0.49, and 0.55 with LERC (Table \ref{table:lercexp}). This indicated human judgments of short answer equivalence correlate well with LERC, making it a viable answer equivalence metric to use as supervision.

\begin{table}[h]

  \centering
  \begin{tabular}{@{}cc@{}}
    \toprule
    Annotators  & Kendall $\tau$  \\
    \midrule
    1 / LERC & 0.55  \\
    2 / LERC & 0.49 \\
    3 / LERC & 0.55 \\
    1 / 2 & 0.38 \\
    2 / 3 & 0.40 \\
    1 / 3 & 0.40 \\
    \bottomrule
  \end{tabular}
  \caption{Inter-annotator agreement across 20 examples and 3 annotators. 2/3 refers to the agreement between annotators 2 and 3}
  \label{table:lercexp}
\end{table}

\section{ClaimDecomp Example}
\label{section:appendix:claimdecomp}

See Table \ref{tab:ClaimDecomp_example}

\section{GPT-4 Prompts}
\label{section:appendix:relprompt}

\begin{prompt}[title={\thetcbcounter{} Relevance Prompt}, label=relevanceprompt]

You will be given a claim, a question about the claim, and a passage. Your job is to check whether the passage contains information that supports an answer to the question. You will only output "Yes" or "No". 

\promptsubsection{Claim}
Hunter Biden had no experience in Ukraine or in the energy sector when he joined the board of Burisma.

\promptsubsection{Question}
Did Hunter Biden have any experience in the energy sector at the time he joined the board of the Burisma energy company in 2014? 

\promptsubsection{Passage}
Hunter Biden , Burisma , Ukraine , and Joe Biden explained - Vox And during the bulk of this troubled period in Hunter ’ s life , he was fortuitously on the board of a Ukrainian energy company... 

\end{prompt} 

\begin{prompt}[title={\thetcbcounter{} Synthetic Data Generation Prompt}, label=syntheticprompt]

You will be provided with a claim and a question about the claim. Your job is to generate two evidence paragraphs: \\
\promptsubsection{(1) Positive} A paragraph that supports an indirect answer to the claim. It requires a reasoning hop to arrive at the answer. You can make up the answer to the question, but it should only come with a reasoning step. \\
\promptsubsection{(2) Hard Negative} A paragraph that looks highly relevant to the claim/question, but doesn't actually support an answer
Neither paragraph can use "claim" or "question" - they must stand alone and mimic the style of real evidence documents found on the web. \\

Here are some examples: \\
\promptsubsection{Claim} Former President Donald Trump who lost the popular vote by 3 million has nominated a full third of The United Supreme Court, as of 13th October 2020. \\
\promptsubsection{Question} How many federal judges did Trump nominate? \\
\promptsubsection{Positive} Two weeks ago in October Trump nominated multiple members of the Supreme Court. He started by nominating John Jacobs and Patricia McConnell, both of whom have supported Republican policies for many years. He made these judicial appointments despite mass disagreement, highlighting his goal to secure conservative ideals in the judiciary. Last week, he also appointed Max Dermott, making him the third Supreme Court justice nominated by Trump. \\
\promptsubsection{Hard Negative} Former President Trump nominated highly conservative Supreme Court justices back in October of 2020. His appointments were largely composed of conservative Republicans with long standing connections to Trump. He made these appointments in accordance with mass public support. \\
\promptsubsection{Explanation} The reasoning step in the positive parargaph is to realize "third of the Supreme court" means 3 out of 9 judges. The positive paragraph correctly lists 3 judges (John Jacobs, Patricia McConnell, and Max Dermott). The hard negative paragraph discusses his appointments but offers no information on how many judges he appointed. \\

Here is another example: \\
\promptsubsection{Claim} Anthony Fauci the NIAID director is a democrat.\\
\promptsubsection{Question} Is Anthony Fauci the NIAID director registered with a political party? \\
\promptsubsection{Positive} Two weeks ago, a new rule was passed in the NIAID which bans any director from holding political affiliations. In fact, it's even stricter than this - the same rule states no NIAID director is allowed to even register with a political party or participate in elections.\\
\promptsubsection{Hard Negative} Anthony Fauci has maintained a long standing relationship with Democratic presidential nominee Jacob Wallace. They were childhood friends who grew up together, and Fauci has also openly supported some of Wallace's policies. However, Fauci is historically known to stray away from politics and media. \\
\promptsubsection{Explanation} The reasoning step in the positive paragraph is to realize NIAID directors cannot register to political parties. Anthony Fauci is an NIAID director according to the claim, therefore he cannot be registered with a political party. The hard negative paragraph mentions his friendship with a Democratic presidential nominee, but this does not imply he is a registered Democrat. \\

Here is one final, slightly harder example: \\
\promptsubsection{Claim} Robert E. Lee, commander of the Confederate States Army during the American Civil War, was not a slave owner.\\
\promptsubsection{Question} Was Robert E. Lee a slave owner?\\
\promptsubsection{Positive} Many commanders during the Civil War era managed and inherited slaves through their family estates. Robert E. Lee was the commander for the Confederate States Army during the Civil War, and the Confederate states were in support of slavery.\\
\promptsubsection{Hard Negative} Commander Robert E. Lee led the Confederate States Army during the American Civil War. In the South, many slaves were forced to fight in the army under Robert E. Lee against the Union states. Slaves as soldiers were given poor equipment and placed on the front lines of defense.\\
\promptsubsection{Explanation} The reasoning step in the positive paragraph is to realize many commanders inherited slaves, and Robert E. Lee was a commander. Therefore it is likely that he might have also had slaves. The hard negative paragraph discusses the role of slaves in the war, but doesn't contain information on whether Robert E. Lee personally owning slaves. Notice even the positive paragraph doesn't contain a direct answer, but it is still more relevant to the question than the hard negative.\\

Now, please generate a positive and hard negative paragraph with an explanation for the following claim/question pair:

\promptsubsection{Claim}
Hunter Biden had no experience in Ukraine or in the energy sector when he joined the board of Burisma.

\promptsubsection{Question}
Did Hunter Biden have any experience in the energy sector at the time he joined the board of the Burisma energy company in 2014? 

\end{prompt}

\begin{prompt}[title={\thetcbcounter{} QA Prompt}, label=gptansprompt]

As a professional fact-checker, your task is to ONLY use the passage to answer the following question about the claim. Keep your answer short (only 1-2 sentences) 

\promptsubsection{Passage} Hunter Biden , Burisma , Ukraine , and Joe Biden explained - Vox And during the bulk of this troubled period in Hunter ’ s life , he was fortuitously on the board of a Ukrainian energy company... 

\promptsubsection{Claim} Hunter Biden had no experience in Ukraine or in the energy sector when he joined the board of Burisma. 

\promptsubsection{Question} Did Hunter Biden have any experience in the energy sector at the time he joined the board of the Burisma energy company in 2014? 

\end{prompt}

\begin{prompt}[title={\thetcbcounter{} FEVER Veracity Prompt}, label=feververacityprompt]

As a professional fact-checker, your task is to use the following claim and evidence document to determine the veracity of the claim. You must ONLY respond with either SUPPORTS, REFUTES, or NOT ENOUGH INFO 

\promptsubsection{Claim} Great white sharks do not prefer dolphins as prey.

\promptsubsection{Passage} Do Sharks Eat Dolphins ? [ Explained ] - Ocean Fauna Did you know that sharks are often considered the ocean ’ s top predators ? Well , here ’ s an interesting twist : killer whales , which are actually a type of dolphin , are the ultimate predators that can effortlessly take down a shark . But what about other dolphin species ? Do sharks eat dolphins ? Not all sharks eat dolphins , but some species do feed on them . Great whites , tiger sharks , and bull sharks are among the ones that go for it . In this article , I will discuss the types of dolphins that sharks typically consume and how they do it . Are Dolphins Prey Items for Sharks ? Yes , dolphins can indeed be prey items for certain species of sharks . While not all sharks view dolphins as a potential meal , some of the larger and more predatory species , such as Great whites , tiger sharks , and bull sharks , have been known to feed on them . Young and sick dolphins are particularly vulnerable to shark predation , as they have less mobility and are easier to catch . Sharks typically hunt

Now please respond with one of SUPPORTS, REFUTES, or NOT ENOUGH INFO
\end{prompt}

\section{Synthetic Data Examples}

\subsection{Generated Example}
\label{section:appendix:syntheticgenerated}

See Table~\ref{table:synthetic_ex}.

\begin{table*}[h]
\small
  \centering
  \begin{tabular}{p{0.1\textwidth}|p{0.8\textwidth}}
    \toprule
    Claim & The Obama-Biden administration secretly launched a surveillance operation on the Trump campaign. \\
    \midrule
    Question & Was there a surveillance operation on the Trump campaign? \\
    \midrule
    Positive & In 2016, it was revealed that the FBI had launched an investigation into potential collusion between the Trump campaign and Russian operatives. This investigation, known as "Crossfire Hurricane," involved surveillance measures such as wiretaps and informants to gather intelligence. Although the operation was conducted by the FBI, it was during the Obama-Biden administration, raising questions about the extent of their involvement and oversight in the surveillance activities. \\
    \midrule
    Hard Negative & During the 2016 presidential campaign, the Trump campaign faced numerous allegations of collusion with foreign entities. These allegations led to a series of public investigations and hearings, where various campaign officials were called to testify. The media extensively covered these events, and the public was made aware of the ongoing scrutiny faced by the Trump campaign. However, no direct evidence of a secret surveillance operation by the Obama-Biden administration was presented during these hearings. \\
    \midrule
    Alt Q1 & What were the objectives or reasons behind the alleged surveillance operation on the Trump campaign? \\
    \midrule
    Alt Q1 Negative ($d_1^-$) & The alleged surveillance operation on the Trump campaign, purportedly initiated by the Obama-Biden administration, was reportedly driven by concerns over potential foreign interference in the 2016 U.S. presidential election. Intelligence agencies had gathered evidence suggesting that Russian operatives were attempting to influence the election's outcome, raising alarms about possible collusion between the Trump campaign and Russian entities. The objective was to safeguard the integrity of the electoral process and ensure that no foreign power could unduly sway the democratic proceedings. Additionally, the surveillance aimed to uncover any illicit activities that could compromise national security. These actions were framed within the broader context of protecting American democratic institutions from external threats.\\
    \midrule
    Explanation & The positive paragraph mentions the FBI's "Crossfire Hurricane" investigation, which involved surveillance measures and occurred during the Obama-Biden administration. This requires the reader to make the reasoning hop that the administration might have had some level of involvement or oversight. The hard negative paragraph discusses public investigations and hearings related to the Trump campaign but does not address the existence of a secret surveillance operation by the Obama-Biden administration. \\
    \bottomrule
  \end{tabular}
  \caption{Example of a synthetic example generated from our procedure. The explanation indicates the reasoning hop required to surface the positive paragraph, as well as the complexity of the hard negative.}
  \label{table:synthetic_ex}
\end{table*}

\subsection{Human Written Example}
\label{section:appendix:synthetichuman}

See Table~\ref{table:human_exs}.

\begin{table*}[h]
\small
  
  \begin{tabular}{p{0.3\textwidth}|p{0.6\textwidth}}
    \toprule
    Claim & Former President Donald Trump who lost the popular vote by 3 million has nominated a full third of The United Supreme Court, as of 13th October 2020. \\
    \midrule
    Question & How many federal judges did Trump nominate? \\
    \midrule
    Positive & Two weeks ago in October Trump nominated multiple members of the Supreme Court. He started by nominating John Jacobs and Patricia McConnell, both of whom have supported Republican policies for many years. He made these judicial appointments despite mass disagreement, highlighting his goal to secure conservative ideals in the judiciary. Last week, he also appointed Max Dermott, making him the third Supreme Court justice nominated by Trump. \\
    \midrule
    Hard Negative & Former President Trump nominated highly conservative Supreme Court justices back in October of 2020. His appointments were largely composed of conservative Republicans with long standing connections to Trump. He made these appointments in accordance with mass public support. \\
    \midrule
    Explanation & The reasoning step in the positive parargaph is to realize ``third of the Supreme court'' means 3 out of 9 judges. The positive paragraph lists 3 judges (John Jacobs, Patricia McConnell, and Max Dermott). The hard negative paragraph discusses his appointments but offers no information on how many judges he appointed, which is what the question is asking. \\
    \bottomrule
  \end{tabular}
  \caption{Example of a human annotated positive and hard negative example. }
  \label{table:human_exs}
\end{table*}

\begin{table*}[t]
\small
  
  \textbf{Claim:} Charles Schumer stated on May 13, 2015 in remarks to reporters: "It is simply a fact that insufficient funding for Amtrak has delayed the installation" of a positive train control safety system. "To deny a connection between the (derailment in Philadelphia) and underfunding Amtrak is to deny reality."

  \textbf{Question:} Is there a connection between the derailment in Phil. and underfunding Amtrak?

  \centering
  \begin{tabular}{p{0.48\linewidth} | p{0.48\linewidth}}
    \toprule
    Contriever & distill (gold) + LERC \\
    \midrule
    Latest safety technology wasn ’ t fully installed at site of deadly Amtrak derailment south of Seattle - CBS News approached sharp curves at more than double the speed limit . A Metro-North train crashed in New York City in 2013 , killing four people , when an engineer with sleep apnea dozed off . An Amtrak train crashed in Philadelphia in 2015 , killing eight people , when investigators say the engineer was distracted by radio traffic and lost his bearings . Positive train control was installed on 23 percent of the nation 's passenger route miles and 37 percent of freight route miles as of July , the last time the Federal Railroad Administration updated its online tracker for the technology . It is activated on the tracks Amtrak owns along the Northeast Corridor , from Boston to Washington , D.C. , and on Amtrak 's Michigan line . Many of its locomotives are equipped for positive train control . Throughout the rest of the country , Amtrak operates on track owned by freight carriers and other entities that have made varying progress on installing the technology . The new \$ 180.7 million route was designed to speed up service by removing passenger trains from a route along Puget Sound that 's bogged down by curves , single-track & Explainer : Positive Train Control and the Amtrak 188 derailment - WHYY Positive Train Control ( PTC ) would have prevented Amtrak 188 from derailing Tuesday , National Transportation Safety Board lead investigator Robert Sumwalt said this week . Amtrak was intending to install the safety system on the Northeast Corridor by the end of year , pursuant to an unfunded congressional mandate under the Rail Safety Improvement Act . \textbf{Amtrak CEO Joseph Boardman has said that , with more funding , Amtrak could have implemented PTC sooner} . The stretch of the Northeast Corridor where the derailment occurred currently uses an older system , Automatic Train Control ( ATC ) . On the southbound route , the ATC “ enforces ” — automatically stops — a train if it is travelling above 45 miles per hour . The northbound side , where Amtrak 188 was travelling , does not . PTC is essentially a smarter version of ATC . Whereas ATC relies on the signal and fixed block system trains have operated on for decades , PTC uses a GPS and radio technology to locate where the trains are along the track . ATC only knows when a train trips a signal wire entering into another large stretch between interlockings or \\
    \midrule
    \textbf{Answer from GPT:} The passage does not provide information on Amtrak's funding levels or directly link underfunding to the derailment in Philadelphia. & \textbf{Answer from GPT:} Yes, according to Amtrak CEO Joseph Boardman, more funding could have allowed Amtrak to implement PTC sooner, which would have prevented the derailment \\
    \bottomrule
  \end{tabular}
  \caption{Comparison of top-1 document on an example from ClaimDecomp between unfinetuned Contriever (left) and CFR model (right). The finetuned retriever is able to surface a document about funding, which is the key aspect the question is targeting.}\label{tab:ClaimDecomp_example}
\end{table*}

\end{document}